\documentclass[conference]{IEEEtran}
\IEEEoverridecommandlockouts

\usepackage{cite}
\usepackage{amsmath,amssymb,amsfonts}
\usepackage{algorithmic}
\usepackage{graphicx}
\usepackage{textcomp}
\usepackage{booktabs}
\usepackage{multirow}
\usepackage{makecell}
\usepackage{url}
\usepackage{array}
\def\BibTeX{{\rm B\kern-.05em{\sc i\kern-.025em b}\kern-.08em
    T\kern-.1667em\lower.7ex\hbox{E}\kern-.125emX}}

\newcommand{\best}[1]{\textbf{#1}}

\begin{document}

\title{Data Scale, Not Latency, Shapes Cross-Lingual Encoder Transfer in Streaming ASR}

\author{
\IEEEauthorblockN{Nenad Banfic}
\IEEEauthorblockA{\textit{CoreAI, Microsoft}}
}

\maketitle

\begin{abstract}
Adapting a streaming speech recognition model to a new language
requires choosing between two plausible warm starts: a multilingual
(ML) encoder or an English-only (EN) encoder.  The common intuition
is that the multilingual encoder should help most at low data, but
it is unclear how long that advantage persists, whether tight
streaming latency amplifies it, and whether it survives deployment
quantization.  We answer these questions with a controlled sweep of
a 0.6\,B-parameter cache-aware FastConformer transducer across eight
European languages, up to five target-language data scales
(100\,h to 2500\,h), three streaming tiers plus offline decoding,
and up to four public test sets.

The main result is that multilingual initialization is a
\emph{data-limited} advantage, not a \emph{latency-limited} one.  On
FLEURS at 160\,ms, the mean EN--ML word error rate (WER) gap
falls from $+4.21$ percentage points (pp) at 100\,h to $+0.20$\,pp at 2500\,h; a
power-law fit summarizes this decay, with each doubling of
target-language data roughly halving the remaining advantage.
Across the three streaming tiers, the across-language mean EN--ML
gap is approximately stable at each scale from 100 to 1000\,h, and
is near zero by 2500\,h.  Finally, 4-bit weight-only encoder
quantization at the matched 560\,ms streaming tier reduces the
encoder footprint by about $3{\times}$, with an average FLEURS
WER increase of about $0.5$\,pp.  The
resulting guideline is simple: use multilingual initialization in
low-data regimes, treat the choice as effectively irrelevant at
large data, and make latency and quantization decisions
independently.
\end{abstract}


\section{Introduction}

Production automatic speech recognition (ASR) increasingly requires
\emph{streaming} operation: emitting partial hypotheses with bounded
look-ahead so that voice assistants, live captioning, and real-time
translation feel responsive.  Modern cache-aware Conformer
architectures~\cite{gulati2020conformer,rekesh2023fastconformer,noroozi2024stateful}
satisfy this requirement by
constraining the encoder's left and right contexts, maintaining
activation caches across non-overlapping audio chunks, and
supporting multi-lookahead training so that a single model can be
deployed at multiple latency operating points.

Bringing such a model to a new language is, in practice, a
fine-tuning problem.  An organization holding both a
0.6-billion-parameter (0.6\,B) English streaming encoder and a
0.6\,B-parameter multilingual streaming encoder must choose, per
target language and per data budget, which to initialize from.
Prior work on cross-lingual transfer for streaming ASR establishes
that pretraining helps over random
initialization~\cite{joshi2020transfer,pandey2024scalable}, and
multilingual speech foundation models continue to push the state of
the art~\cite{conneau2020xlsr,zhang2023google}.
However, the practically interesting comparison is \emph{between two
viable pretrained encoders} that differ in pretraining language
coverage, not between pretraining and random initialization; a
random-encoder sanity check under the same recipe was not competitive
and is reported in the supplement.  Throughout this paper we use
\emph{multilingual} (ML) for the encoder trained jointly on
multiple languages, and \emph{English-only} (EN) for the encoder
trained on English data only.  Existing transfer studies for
streaming ASR cover one or two target languages
(Hindi~\cite{joshi2020transfer}, French/Italian~\cite{pandey2024scalable}),
do not vary streaming latency at evaluation, and do not include
cache-aware Conformers; multilingual foundation-model work
(XLSR~\cite{conneau2020xlsr}, USM~\cite{zhang2023google})
reports many languages but evaluates
in offline/non-streaming settings rather than across
streaming-latency tiers.

We sweep four axes simultaneously --- target language (eight European
languages spanning Germanic, Romance and Slavic), training data
scale (up to five levels per language, 100\,h to 2500\,h),
streaming-inference latency (three streaming tiers at 160, 560 and
1120\,ms plus an offline batch reference; we reserve
``streaming tier'' for the three throughout, naming offline
explicitly when it is included), and evaluation domain (up to four
public test sets) --- for
both ML and EN initializations.  The eight target languages are
German (DE), Dutch (NL) and Icelandic (IS) from Germanic; Spanish
(ES), French (FR) and Portuguese (PT) from Romance; and Croatian
(HR) and Polish (PL) from Slavic.  The result is, to our knowledge,
the largest published cross-lingual transfer study for a
streaming-trained ASR architecture, spanning seed-reproducibility,
Slavic-pivot and hybrid-encoder runs, plus INT4 quantization
sweeps over the main grid.

The headline finding is that multilingual initialization is a
\emph{data-limited} advantage, not a \emph{latency-limited} one: across
the eight languages the mean EN--ML FLEURS gap closes monotonically
with target-language hours and is well summarized by a power law
(Sec.~\ref{sec:hours}), while the across-language mean gap is
approximately stable across the three streaming-latency tiers at
every data scale where it is non-trivial (Sec.~\ref{sec:latency}).
4-bit weight-only encoder quantization at the matched 560\,ms
tier costs roughly half a WER point on FLEURS
(Sec.~\ref{sec:quant}), small compared to the EN--ML gap at low
data.  Per-language and per-corpus breakdowns
(Sec.~\ref{sec:per_lang}), a Slavic-pivot warm start, and a
hybrid-encoder ablation that localizes the transferable signal to
the upper half of the encoder stack (both in Sec.~\ref{sec:support}),
together with a two-seed reproducibility re-run on all HR/PT cells
(Sec.~\ref{sec:recipe},~\ref{sec:hours}), extend this picture
without changing the headline conclusion.

\section{Related Work}

\textbf{Transfer learning for streaming ASR.}
Joshi et~al.~\cite{joshi2020transfer} studied transfer for a
recurrent neural network transducer
(RNN-T)~\cite{graves2012sequence} to Hindi at 50--1{,}000\,h,
showing that pretraining is most useful at low data and accelerates
convergence.  Pandey et~al.~\cite{pandey2024scalable}
extended this line of work to Emformer on-device models in French and Italian.
Neither study varies streaming latency at evaluation, and both
compare transfer-initialized models against non-pretrained target-language
baselines rather than two pretrained
initializations.

\textbf{Multilingual ASR at scale.}
Multilingual end-to-end models~\cite{kannan2019large,li2021scaling},
large-scale multilingual systems~\cite{zhang2023google}, and
self-supervised representations~\cite{conneau2020xlsr,pratap2023mms}
show that learning across many languages can match or beat
per-language monolingual baselines.  OWLS~\cite{chen2025owls}
derives neural scaling laws across model, data and compute scale,
complementing earlier
acoustic-model scaling laws~\cite{droppo2021scaling}; we instead hold
model size fixed and study how downstream fine-tuning data and latency
interact with encoder initialization.  Hernandez et~al.~\cite{hernandez2021scaling}
showed that, in the low-data regime, the effective data transferred
from text pretraining to code fine-tuning follows a power law in
model size and fine-tuning data size.

\textbf{Cache-aware streaming.}
The stateful Conformer work of Noroozi et~al.~\cite{noroozi2024stateful}
adapts FastConformer to streaming ASR with constrained attention context
and multi-latency training, enabling a single model
to support multiple inference latencies.  Related cascaded-encoder
architectures~\cite{narayanan2021cascaded} pursue the complementary goal
of unifying streaming and non-streaming operation within one model.  To
our knowledge, ours is the first transfer-learning study to treat
streaming-latency tiers as an explicit evaluation axis.

\section{Experimental Setup}
\label{sec:setup}

\subsection{Model and initializations}

We use a 0.6\,B-parameter cache-aware FastConformer RNN-T with 24
encoder layers, $d_{\text{model}}{=}1024$, an 8$\times$ subsampling
pre-encoder, and a long short-term memory (LSTM)-based RNN-T
decoder with joint network~\cite{noroozi2024stateful}, trained with
the NeMo toolkit~\cite{kuchaiev2019nemo}.  The encoder
is trained with multiple \texttt{att\_context\_size} configurations
so it can be deployed at any of the latency tiers in
Table~\ref{tab:latency}.  We compare two initializations:
(i) \textbf{ML} loads all weights from the multilingual
streaming checkpoint described below; and (ii) \textbf{EN} loads
decoder and joint from that same multilingual baseline, but
replaces the encoder layer-for-layer with the English-only
Nemotron streaming encoder~\cite{noroozi2024stateful,nemotronstreamingen}.  Both arms
share the same multilingual byte-pair-encoding (BPE)
tokenizer~\cite{sennrich2016bpe}, so
the comparison isolates the encoder.

\begin{table}[!t]
\centering
\caption{Streaming-inference operating points used in evaluation.
All three streaming tiers share the same 70-frame left context
($\approx 5.6$\,s of audio); each chunk spans the right context
plus the current frame, and the chunk size equals the emission
latency.  The encoder runs at 80\,ms/frame after
8$\times$ subsampling.  \textsc{Offline} decodes the full utterance
in a single pass with the same context caps as the 1120\,ms tier.}
\label{tab:latency}
\setlength{\tabcolsep}{6pt}
\begin{tabular}{lccc}
\toprule
\textbf{Tier} & \textbf{Chunk size} & \textbf{Left ctx} & \textbf{Right ctx} \\
\midrule
160\,ms  & 2 frames                & 70 frames & 1 frame   \\
560\,ms  & 7 frames                & 70 frames & 6 frames  \\
1120\,ms & 14 frames               & 70 frames & 13 frames \\
offline  & --                      & 70 frames & 13 frames \\
\bottomrule
\end{tabular}
\end{table}

\textbf{Multilingual baseline.}
The ML initialization is a multilingual checkpoint,
trained once and then frozen as the starting point for every ML
fine-tuning run.
We start from the same English encoder used as the EN
initialization~\cite{nemotronstreamingen},
\texttt{nvidia/nemotron-speech-streaming-en-0.6b}, replace its
tokenizer and joint network with those of
\texttt{nvidia/parakeet-tdt-0.6b-v3}~\cite{parakeettdtv3} so that the vocabulary covers
the target European languages, and adapt the full model to five
European languages (DE, ES, FR, NL and Italian IT) with standard
RNN-T training.  The corpus is the union of the
public training portions of Common Voice (CV)~\cite{ardila2020commonvoice},
Multilingual LibriSpeech (MLS)~\cite{pratap2020mls},
VoxPopuli (VP)~\cite{wang2021voxpopuli}, CML-TTS~\cite{cmltts}
and crawled YODAS-Granary~\cite{raokoluguri2025granary} subsets,
$\sim 9$\,k hours.  Training uses an
effective batch size of 448 on 7 NVIDIA H100 GPUs.  The downstream
fine-tuning manifests of Sec.~\ref{sec:data} are constructed
independently and verified utterance-disjoint from this corpus;
FLEURS is excluded from both training stages.

\subsection{Target languages and corpora}
\label{sec:data}

We fine-tune on eight European languages spanning three families:
Germanic (DE, NL, IS), Romance (ES, FR, PT) and Slavic (HR, PL).
We further partition them by their representation in the multilingual
pretraining mix:
\begin{itemize}
  \item \textbf{Seen} (DE, ES, FR, NL): well covered by the ML
        pretrain corpus; the comparison against EN init mainly
        quantifies the head-room left.
  \item \textbf{Unseen / under-represented} (PT, HR, PL, IS):
        absent from, or only marginally represented in, the ML
        pretrain mix.  These are the cases that test whether the
        multilingual prior survives outside its training
        distribution.
\end{itemize}

\textbf{Fine-tuning corpora.}  Training data is drawn from the
training splits of the same public corpora as the multilingual
baseline above (CV, MLS, VoxPopuli, CML-TTS and YODAS-Granary).  Two of
our unseen languages have limited coverage in these corpora and
use additional language-specific sources: Croatian draws the bulk
of its hours from the ParlaSpeech-HR parliamentary
corpus~\cite{parlaspeech_hr}, and
Icelandic combines the Althingi Parliamentary Speech
Corpus~\cite{althingi} with the Samr\'{o}mur~\cite{samromur} and
M\'{a}lr\'{o}mur~\cite{malromur} crowdsourced read-speech corpora.
For each language we
construct nested manifests at 100, 250, 500, 1000 and 2500\,h
where sufficient data exists; HR, IS and PL stop at 1000\,h.

\textbf{Evaluation sets.}  We evaluate on the official test splits
of CV, MLS, VoxPopuli and FLEURS~\cite{fleurs2022}.  The first three
are \emph{in-domain} (their training splits feed the fine-tuning
manifests); FLEURS is \emph{out-of-domain}, held out from
fine-tuning at every scale, and is our headline set.  All evaluations report WER over
hypotheses and references normalized with the
\texttt{BasicMultilingualTextNormalizer} from the Open ASR
Leaderboard~\cite{openasrleaderboard}.  For each per-cell
$\Delta$ on FLEURS we run a per-utterance paired
bootstrap~\cite{bisaniney2004} ($B{=}1000$) and mark $\Delta$ with
$^{\ast}$ when its 95\% CI excludes 0 (Table~\ref{tab:main160}).

\subsection{Training recipe and seeds}
\label{sec:recipe}

All fine-tuning runs use the same recipe: RNN-T loss with FastEmit
regularization~\cite{yu2021fastemit} ($\lambda{=}0.005$),
AdamW~\cite{loshchilov2019adamw}, learning rate
$1{\times}10^{-4}$ with cosine decay to $10^{-6}$ over the first 25
epochs of a 30-epoch schedule,
effective batch size 48 on NVIDIA H100 GPUs in bf16 mixed precision, weight decay $10^{-3}$,
decayed SpecAugment~\cite{park2019specaugment} ($f{=}2{\times}27$, $t{=}10{\times}0.05$),
early stopping with patience 8.  Checkpoint selection and early
stopping use WER on the target-language FLEURS dev split.  This
recipe was fixed by a small pilot sweep and then frozen for every
(lang, hours, initialization) cell.
Training samples encoder context uniformly over $[70,13]$, $[70,6]$, $[70,1]$, and $[70,0]$.

Every (init, hours) cell of HR and PT was trained both with the
default training seed (42) and with seed 45; additionally, the
100\,h cell of each of the other six languages was re-trained with
seed 45 for both initializations.

\section{Main Results}
\label{sec:main}

\subsection{ML's advantage closes with hours}
\label{sec:hours}

\begin{table*}[!t]
\centering
\caption{FLEURS WER (\%) at the 160\,ms streaming-latency tier.  $\Delta = \mathrm{WER}_{\mathrm{EN}} - \mathrm{WER}_{\mathrm{ML}}$ (positive: ML init better).  $^{\ast}$ marks cells where the paired per-utterance bootstrap 95\% CI for $\Delta$ excludes 0 ($B{=}1000$).  Best per (lang, hours) in \best{bold}.  ``--'': data not available.  The bottom \textnormal{\textit{mean}} row reports the unweighted across-language macro mean ($K{=}8$ for 100--1000\,h, $K{=}5$ for 2500\,h).  $\Delta$ is computed at full precision, so it can differ from the difference of the rounded WERs by $0.01$.}
\label{tab:main160}
\renewcommand{\arraystretch}{1.05}
\setlength{\tabcolsep}{3pt}
\small
\begin{tabular}{l ccc ccc ccc ccc ccc}
\toprule
& \multicolumn{3}{c}{\textbf{100\,h}} & \multicolumn{3}{c}{\textbf{250\,h}} & \multicolumn{3}{c}{\textbf{500\,h}} & \multicolumn{3}{c}{\textbf{1000\,h}} & \multicolumn{3}{c}{\textbf{2500\,h}} \\
\cmidrule(lr){2-4}\cmidrule(lr){5-7}\cmidrule(lr){8-10}\cmidrule(lr){11-13}\cmidrule(lr){14-16}
\textbf{Lang} & \textbf{ML} & \textbf{EN} & $\boldsymbol{\Delta}$ & \textbf{ML} & \textbf{EN} & $\boldsymbol{\Delta}$ & \textbf{ML} & \textbf{EN} & $\boldsymbol{\Delta}$ & \textbf{ML} & \textbf{EN} & $\boldsymbol{\Delta}$ & \textbf{ML} & \textbf{EN} & $\boldsymbol{\Delta}$ \\
\midrule
DE & \best{16.57} & 21.73 & +5.16$^{\ast}$ & \best{14.93} & 16.72 & +1.78$^{\ast}$ & \best{13.04} & 13.98 & +0.94$^{\ast}$ & \best{13.33} & 13.89 & +0.56$^{\ast}$ & 11.37 & \best{11.07} & -0.30 \\
ES & \best{11.12} & 16.43 & +5.31$^{\ast}$ & \best{10.05} & 12.08 & +2.03$^{\ast}$ & \best{8.70} & 9.90 & +1.20$^{\ast}$ & \best{8.97} & 9.76 & +0.79$^{\ast}$ & \best{7.38} & 7.80 & +0.43$^{\ast}$ \\
FR & \best{19.51} & 25.38 & +5.87$^{\ast}$ & \best{16.84} & 19.87 & +3.02$^{\ast}$ & \best{15.55} & 16.81 & +1.26$^{\ast}$ & \best{15.49} & 16.47 & +0.97$^{\ast}$ & \best{13.75} & 13.81 & +0.06 \\
HR & \best{36.36} & 38.29 & +1.92$^{\ast}$ & \best{28.94} & 30.08 & +1.14$^{\ast}$ & \best{24.91} & 25.50 & +0.59 & \best{25.03} & 25.19 & +0.16 & -- & -- & -- \\
IS & \best{26.23} & 27.88 & +1.65 & \best{22.85} & 23.04 & +0.19 & 20.62 & \best{20.33} & -0.29 & 19.26 & \best{19.07} & -0.19 & -- & -- & -- \\
NL & \best{23.00} & 30.80 & +7.80$^{\ast}$ & \best{20.94} & 23.81 & +2.87$^{\ast}$ & \best{18.83} & 20.08 & +1.25$^{\ast}$ & \best{17.85} & 19.23 & +1.38$^{\ast}$ & \best{15.04} & 15.95 & +0.91$^{\ast}$ \\
PL & \best{35.92} & 37.42 & +1.50$^{\ast}$ & \best{27.28} & 27.78 & +0.50 & \best{21.39} & 22.33 & +0.93$^{\ast}$ & \best{21.73} & 22.48 & +0.75$^{\ast}$ & -- & -- & -- \\
PT & \best{15.27} & 19.74 & +4.48$^{\ast}$ & \best{12.45} & 14.52 & +2.06$^{\ast}$ & \best{11.78} & 12.18 & +0.40 & \best{12.05} & 12.13 & +0.08 & 10.24 & \best{10.17} & -0.07 \\
\midrule
\textit{mean} & 23.00 & 27.21 & $+4.21$ & 19.29 & 20.99 & $+1.70$ & 16.85 & 17.64 & $+0.79$ & 16.71 & 17.28 & $+0.56$ & 11.56 & 11.76 & $+0.20$ \\
\bottomrule
\end{tabular}
\end{table*}

Table~\ref{tab:main160} reports FLEURS WER at the most demanding
latency tier (160\,ms) for both initializations across all
available (lang, hours) cells, together with paired-bootstrap
significance markers.  The
mean $\Delta$ drops monotonically from
$+4.21$\,pp at 100\,h ($n{=}8$) to
$+0.79$\,pp at 500\,h,
$+0.56$\,pp at 1000\,h and
$+0.20$\,pp at 2500\,h ($n{=}5$).  The decay is monotone within each language
group: seen languages average
$+6.03 \to +1.16 \to +0.93$\,pp across 100/500/1000\,h, while the
four unseen languages average
$+2.39 \to +0.41 \to +0.20$\,pp over the same scales --- a smaller
absolute advantage that disappears into the noise by 1000\,h.  At 2500\,h no per-language $|\Delta|$
exceeds 1\,pp and several are not statistically distinguishable from
zero under the paired bootstrap.

\begin{figure*}[!t]
\centering
\includegraphics[width=\textwidth]{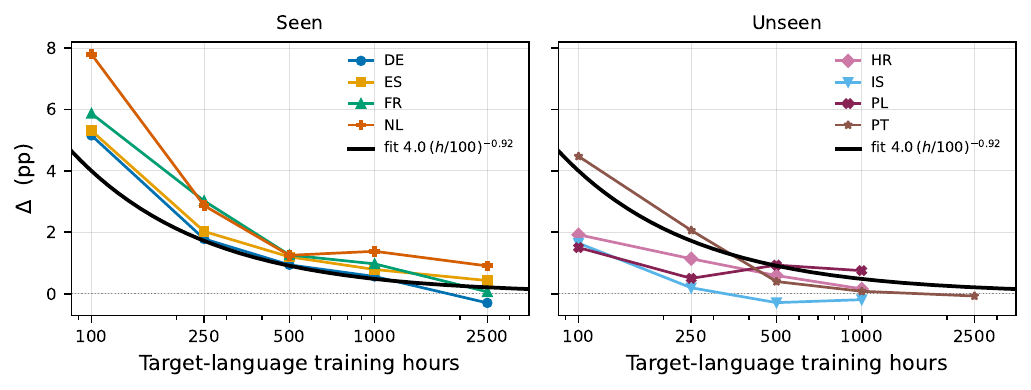}
\caption{Per-language $\mathrm{EN}-\mathrm{ML}$ FLEURS gap
$\Delta_\ell(h)$ (pp) at 160\,ms versus training hours $h$, split into
\emph{seen} (DE, ES, FR, NL; left) and \emph{unseen} (HR, IS, PL, PT;
right) languages.  Black curve: the power-law fit
$\bar\Delta(h){\approx}4\,(h/100)^{-0.92}$ \eqref{eq:scaling_value}
to the mean over all eight languages
($R^{2}{=}0.99$), drawn identically in both panels for reference.}
\label{fig:scaling_curve}
\end{figure*}

\textbf{Transfer-gap decay.}  A natural practitioner question is:
given $h$ hours of target-language data, how much initialization
advantage remains from a multilingual checkpoint?  Writing
$\bar\Delta(h)$ for the across-language mean EN--ML WER gap (pp) at
$h$ hours and $h_0{=}100$\,h for the lowest scale, we summarize the
decay with a power-law fit
\begin{equation}
  \bar\Delta(h)=\bar\Delta(h_0)\,(h/h_0)^{-\beta_{\mathrm{TG}}},
  \label{eq:scaling}
\end{equation}
whose exponent $\beta_{\mathrm{TG}}$ we call the \emph{transfer-gap
exponent}: it sets how fast the EN--ML advantage decays with
target-language data.
Fitting on FLEURS at
160\,ms gives
\begin{equation}
  \bar\Delta(h)\approx 4\,(h/100)^{-0.92},
  \quad R^{2}=0.99
  \label{eq:scaling_value}
\end{equation}
(Fig.~\ref{fig:scaling_curve}).  Power-law behavior is familiar from
the transfer-learning scaling-laws literature~\cite{hernandez2021scaling},
although Hernandez et\,al.\ model effective transferred data rather
than the WER gap between two pretrained initializations.  With $\hat\beta_{\mathrm{TG}}{\approx}0.92$, doubling
the amount of target-language data multiplies the remaining ML-init
advantage by $2^{-0.92}{\approx}0.53$, i.e., it roughly halves the
gap.

A language-resampling bootstrap ($B{=}10000$) at 160\,ms gives a 95\% CI of
$[0.59,\,1.52]$ for $\hat\beta_{\mathrm{TG}}$, with fewer than
1\% of resamples yielding $\hat\beta_{\mathrm{TG}}{<}0.5$.
Refitting the same form on the three in-domain test sets gives
$\hat\beta_{\mathrm{TG}}{\in}\{0.79~(\text{VP}),\,0.97~(\text{MLS}),\,1.09~(\text{CV})\}$
($R^{2}{\geq}0.93$), and applying the FLEURS@160\,ms fit unchanged to
the other latency tiers predicts each per-tier mean gap to within
$\sim 0.5$\,pp on average.  One fixed exponent thus fits the
across-language mean across all four test sets and four latency
settings with no retuning, and is robust: dropping the 2500\,h point
gives $\hat\beta_{\mathrm{TG}}{=}0.90$, and separate seen/unseen fits
give $0.92$ and $1.09$ ($R^{2}{\geq}0.98$); only per-language fits
scatter more widely.

\textbf{Seed stability.}  Re-running each cell with a second seed,
the mean absolute seed-to-seed FLEURS WER difference at 560\,ms is
$0.21$\,pp across the HR/PT cells and $0.44$\,pp across the 100\,h
cells of all eight languages --- well below the 100\,h EN--ML gaps
at the same tier (up to $+7.43$\,pp for NL).  The EN--ML gap keeps
the same sign at both seeds for all eight languages.

\begin{figure}[!t]
\centering
\includegraphics[width=\columnwidth]{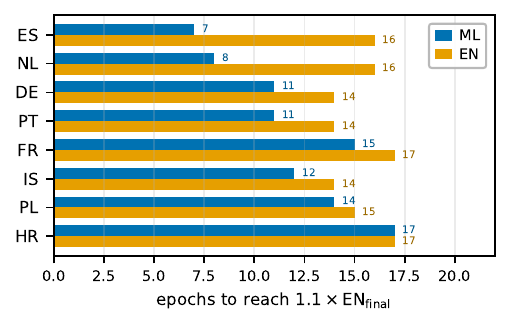}
\caption{1000\,h fine-tuning convergence speed: epochs each arm
needs to reach $1.1{\times}\mathrm{EN}_{\mathrm{final}}$, a WER
target reachable by both arms in every language.}
\label{fig:convergence_1000h}
\end{figure}

\textbf{Convergence speed.}  The multilingual prior also accelerates
fine-tuning: at 1000\,h, ML reaches the EN arm's final
running-best validation WER ($\mathrm{EN}_{\mathrm{final}}$) 4--11 epochs sooner in 6/8 languages,
and reaches a looser $1.1{\times}\mathrm{EN}_{\mathrm{final}}$
target (achievable by both arms in every language) a mean of
$3.5$ epochs sooner (Fig.~\ref{fig:convergence_1000h}).

\textbf{Single-language extension to 5000\,h.}  Since the EN--ML
gap is already near zero on average at 2500\,h, we extended only a
single ES ML-init checkpoint to 5000\,h.  At 560\,ms streaming it
gains $\sim 0.7$--$1.4$\,pp on every ES test set over the 2500\,h
cell and stays competitive with the contemporaneous NVIDIA
Nemotron-3.5-ASR-Streaming-0.6B
checkpoint~\cite{nvidia2026nemotron35streaming} (same size and
streaming family~\cite{noroozi2024stateful}, but multilingual) to
within $1$\,pp on three of the four splits and $1.02$\,pp on
VoxPopuli --- so absolute WER keeps improving past the point where
the initialization gap has closed.

\subsection{The EN--ML gap is latency-invariant}
\label{sec:latency}

\begin{table}[!t]
\centering
\caption{Unweighted across-language (macro) mean $\mathrm{EN}-\mathrm{ML}$ gap (pp) on FLEURS at each streaming-latency tier (and offline batch decoding) versus target-language data scale.  Positive: ML init better.  Each cell is the arithmetic mean across the available languages ($K{=}8$ for 100--1000\,h, $K{=}5$ for 2500\,h) of the per-language $\Delta$ (EN$-$ML) at that tier and data scale; the 160\,ms row aggregates the $\Delta$ column of Table~\ref{tab:main160}.}
\label{tab:latency_effect}
\begin{tabular}{lccccc}
\toprule
\textbf{Tier} & \textbf{100\,h} & \textbf{250\,h} & \textbf{500\,h} & \textbf{1000\,h} & \textbf{2500\,h} \\
\midrule
160\,ms  & $+4.21$ & $+1.70$ & $+0.79$ & $+0.56$ & $+0.20$ \\
560\,ms  & $+4.48$ & $+1.77$ & $+0.85$ & $+0.40$ & $+0.05$ \\
1120\,ms & $+4.56$ & $+1.43$ & $+0.96$ & $+0.27$ & $-0.58$ \\
offline  & $+4.19$ & $+1.61$ & $+0.81$ & $+0.45$ & $+0.09$ \\
\bottomrule
\end{tabular}
\end{table}

Multi-latency cache-aware training
makes cross-tier robustness of \emph{absolute} WER plausible, but
does not by itself imply that the \emph{EN--ML transfer gap}
should be latency-stable.  Our three streaming tiers share the
same 70-frame left context and differ only in chunk size
(Table~\ref{tab:latency}), but initialization and chunking can
still interact: at 160\,ms each chunk holds only two encoder
frames, so the model sees less right-context evidence per emission
step.  A priori, a reasonable
hypothesis is therefore that the multilingual prior might matter
\emph{more} at tight latencies.

We quantify cross-tier stability with the \emph{Latency
Sensitivity of Transfer} (LST).  For language $\ell$ at data scale
$h$,
\begin{align}
  \Delta_\tau(\ell, h) &= \mathrm{WER}_{\mathrm{EN},\tau}(\ell, h)
                         - \mathrm{WER}_{\mathrm{ML},\tau}(\ell, h), \label{eq:delta_tau} \\
  \mathrm{LST}(\ell, h) &= \max_\tau \Delta_\tau(\ell, h)
                          - \min_\tau \Delta_\tau(\ell, h), \label{eq:lst}
\end{align}
where $\tau \in \{160,\,560,\,1120\,\mathrm{ms}\}$.  LST is thus
the spread of the EN--ML gap across the three streaming tiers; a
small LST means the transfer advantage is nearly the same at every
latency.  LST and the across-language macro-mean gap $\bar\Delta$
(Table~\ref{tab:latency_effect}) are complementary: the macro-mean
averages over languages \emph{before} comparing tiers, so
opposite-sign per-language shifts can cancel, whereas LST takes each
language's own across-tier spread \emph{before} averaging and so
cannot be hidden by such cancellation.  Offline decoding is excluded
from LST but reported alongside the streaming tiers as a
latency-unconstrained reference.

\begin{figure}[!t]
\centering
\includegraphics[width=\columnwidth]{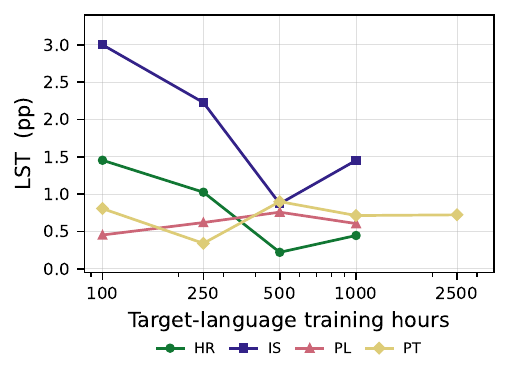}
\caption{Per-language LST~\eqref{eq:lst} on FLEURS versus
target-language training hours for the four \emph{unseen}
languages (HR, IS, PL, PT); computed over
the three streaming tiers (160/560/1120\,ms).  IS shows the largest unseen-language LST values at
100\,h and 250\,h; otherwise most unseen-language cells remain near
or below 1\,pp (HR, IS and PL have no 2500\,h cell).}
\label{fig:lst}
\end{figure}

Averaged over the eight languages, LST~\eqref{eq:lst} is
$1.00$\,pp at 100\,h, $1.13$ at 250\,h, $0.61$ at 500\,h and $0.66$
at 1000\,h: at every data scale the EN--ML gap moves by on the
order of a point as latency varies across the three streaming
tiers.  A paired
per-utterance bootstrap ($B{=}1000$) confirms this is not a
small-sample artifact, with the 95\% CI upper bound on the
across-language mean LST staying $\le 1.6$\,pp at every data scale.
Per-language exceptions exist (Fig.~\ref{fig:lst}), most visibly
IS at 100\,h, so this stability is an aggregate pattern rather than
a property of every (lang, hours) cell.  The
hypothesis that the smaller per-chunk look-ahead at 160\,ms should
amplify the value of the initialization is therefore \emph{not}
supported.

Table~\ref{tab:latency_effect}, which collapses each tier to its
across-language mean gap, tells the same story at the aggregate
level and lets us fold in offline decoding.  Within a fixed data
scale this macro-mean gap $\bar\Delta$, now read per tier, is
approximately constant across the
three streaming tiers --- its range is at most $0.35$\,pp wherever
the gap is non-trivial (100--1000\,h) --- and at 2500\,h all four
values (the three streaming tiers and offline) lie within $1$\,pp
of zero, as expected once the gap has effectively closed.  Offline
tracks the streaming-tier mean to within $0.23$\,pp at every data
scale, and the absolute streaming penalty itself (streaming WER
minus offline WER) is small and roughly equal for ML and EN init:
$\sim 1.1$--$1.2$\,pp at 160\,ms across 100--1000\,h for both
arms.

\subsection{Per-language pattern: the unseen four}
\label{sec:per_lang}

Per-language $\Delta$ values group cleanly by representation in
the multilingual pretraining mix.  The largest 100\,h gains accrue
to the seen languages: NL ($+7.80$\,pp), FR ($+5.87$\,pp),
ES ($+5.31$\,pp), DE ($+5.16$\,pp); seen-group mean $+6.03$\,pp.
The four unseen languages show much smaller gains: PT
($+4.48$\,pp), HR ($+1.92$\,pp), IS ($+1.65$\,pp), PL
($+1.50$\,pp); group mean $+2.39$\,pp.  Both groups' gaps shrink
sharply by 1000\,h: the unseen group as a whole is statistically
indistinguishable from EN init (95\% bootstrap CI on the group-mean
$\Delta$ contains zero), with HR, IS and PT individually
non-significant and only PL retaining a small but significant gap
($+0.75$\,pp).  IS has by far the smallest FLEURS test set
(46 utterances), giving it the widest bootstrap interval and the
only non-significant 100\,h $\Delta$ of the eight languages; we
treat its 100\,h gap as a low-confidence signal.  The unseen group still benefits
modestly from the multilingual prior despite the target language
being out of the pretraining distribution: a hybrid-encoder
ablation (Sec.~\ref{sec:hybrid}) localizes this transferable
component to the upper half of the encoder.

\subsection{Supporting transfer analyses}
\label{sec:support}

\textbf{Slavic pivot: HR fine-tuned from PL.}\label{sec:from_pl}
A cheaper alternative to rebuilding a multilingual pretraining mix is
to warm-start from a model already fine-tuned on a related language.
We test this for HR by pivoting through the PL$_{\text{ML}}$
1000\,h checkpoint, the only same-family Slavic pivot available
among our targets.  The pivot does not improve over direct ML
initialization: on FLEURS it is $+0.56$ to $+1.05$\,pp worse
across the four scales, significant at three of them, while on
VoxPopuli the differences are smaller and mostly within noise,
significant only at 250\,h, where the pivot is instead $1.34$\,pp
better.  This negative result suggests the broad multilingual
encoder is already at least as useful a prior as this
PL-specialized checkpoint, and that same-family relatedness alone
does not determine the best pivot.

\textbf{Layer-wise localization.}\label{sec:hybrid}
A hybrid-encoder ablation combines layers from the ML and EN
checkpoints while keeping the multilingual decoder, joint and
pre-encoder; we fine-tune on 100\,h of the target language with the
recipe of Sec.~\ref{sec:recipe} and evaluate at 560\,ms streaming
(single seed throughout).  The splice that takes the bottom 12
encoder layers from English (input-proximal layers) but keeps the
top 12 multilingual layers (decoder-proximal layers) matches
full-ML accuracy across \emph{all eight} languages:
across-language mean $-0.07$\,pp on FLEURS, with seen-group mean
$17.08$ vs $17.07$\,\% and unseen-group $27.30$ vs $27.45$\,\%.
Other splices --- keeping only the bottom, middle, or top third of
the encoder from ML (the rest from EN) --- all regress, the
middle-third variant essentially back to the full-EN baseline.  This matches the layer-wise probing
literature on speech encoders~\cite{pasad2021layer,yang2021superb}
and gives direct ASR-side confirmation that the transferable
multilingual-pretraining signal sits in the higher encoder layers,
not the acoustic front-end.  A complementary
joint-re-initialization ablation at the same 100\,h cells (both
encoders intact, the same freshly initialized joint for both arms)
hurts the EN-init arm substantially
more than the ML-init arm, even though the reset joint is
fine-tuned together with its encoder.  This argues against a
simple joint--encoder mismatch explanation for the EN--ML gap and
instead suggests that the ML encoder exposes representations that
are easier for the RNN-T decoder/joint stack to use.  Both
ablations still require the full multilingual checkpoint as a
source, so these are representational rather than deployment
findings.

\section{Encoder Quantization}
\label{sec:quant}

For on-device deployment we additionally quantize the encoder
weights to 4-bit integers using the k-quant weight-only scheme of
ONNX Runtime's \texttt{MatMulNBitsQuantizer} (block size 32,
symmetric)~\cite{onnxruntime}; prior work has
reported that this same INT4 k-quant recipe preserves WER within
$\sim 1$\,pp on a 0.6\,B-parameter English Nemotron streaming-ASR
model~\cite{banfic2026ondevice}, and we apply it unchanged here,
without per-language or per-cell quantization tuning.  Only the
encoder's \texttt{MatMul} weights are quantized; the decoder,
joint network, all activations, and the audio front-end remain
FP32.  Since the encoder holds the large majority of model
parameters, quantizing it captures most of the realizable
footprint reduction.

The INT4 ONNX models are deployed via \texttt{onnxruntime-genai}~\cite{onnxruntimegenai}
at the 560\,ms streaming tier.  To make the quantization
comparison fair we rebuild the FP32 reference at the same tier
from the NeMo checkpoint of every fine-tuned cell.

\begin{table}[!t]
\centering
\caption{INT4 weight-only encoder quantization on FLEURS at 560\,ms streaming latency, against an FP32 NeMo baseline at the same latency.  $\bar\Delta_{\mathrm{Q}} = \mathrm{WER}_{\mathrm{INT4}} - \mathrm{WER}_{\mathrm{FP32}}$ (pp, positive = INT4 worse), averaged across the available languages.  Decoder and joint are kept FP32.}
\label{tab:quant}
\small
\begin{tabular}{lrrrr}
\toprule
\textbf{Hours} & $\bar\Delta_{\mathrm{Q,ML}}$ (pp) & $n_{\mathrm{ML}}$ & $\bar\Delta_{\mathrm{Q,EN}}$ (pp) & $n_{\mathrm{EN}}$ \\
\midrule
100      & $+0.88$ & 8 & $+0.70$ & 8 \\
250      & $+0.34$ & 8 & $+0.65$ & 8 \\
500      & $+0.15$ & 8 & $+0.63$ & 8 \\
1000     & $+0.26$ & 8 & $+0.57$ & 8 \\
2500     & $+0.17$ & 5 & $+0.41$ & 5 \\
\midrule
\textit{pooled} & \multicolumn{4}{c}{$+0.49$ (pp), median $+0.42$, $n{=}74$ cells} \\
\bottomrule
\end{tabular}
\end{table}

Across the 74 matched cells (Table~\ref{tab:quant}) the mean change
is $+0.49$\,pp and the
median is $+0.42$\,pp; 60/74 cells lie within $\pm 1.0$\,pp of
their FP32 counterpart and INT4 is worse than FP32 in 65/74
cells.  The eight cells where INT4 is nominally better (the
remaining cell is an exact tie) are all
small in magnitude (within $0.68$\,pp), comparable to the
seed-to-seed variation reported in Sec.~\ref{sec:recipe}.  The per-cell cost is small but \emph{not zero}: it
buys roughly $3{\times}$
encoder file-size reduction at a typical cost of about half a WER
point on FLEURS, with no systematic dependence on data scale.  A
within-(lang, hours) paired comparison gives a mean
$\bar\Delta_{\mathrm{Q,EN}}-\bar\Delta_{\mathrm{Q,ML}}{=}+0.23$\,pp
across the 37 paired cells: the EN arm is more often costlier to
quantize than the ML arm, but the effect is inconsistent across
cells, so we treat this as a weak trend rather than a robust effect.

\section{Conclusion}

We presented a systematic study of how multilingual versus
English-only encoder initialization interacts with target-language
data scale, streaming latency and target language for a
cache-aware streaming ASR architecture, with particular attention to the four
languages (PT, HR, PL, IS) outside the multilingual pretraining
mix.  Across eight languages the multilingual advantage closes
monotonically with data and is approximately latency-invariant:
the across-language mean gap varies by $\le 0.35$\,pp across the
three streaming tiers wherever it is non-trivial, and the latency
sensitivity of transfer (LST), averaged over languages, remains
small, ranging from $0.61$ to
$1.13$\,pp across the 100--1000\,h scales, with the largest value
at 250\,h driven by a small number of language-specific outliers.
A power law with exponent
$\hat\beta_{\mathrm{TG}}{\approx}0.92$ summarizes the gap decay,
and INT4 weight-only encoder quantization at 560\,ms shifts FLEURS
WER by $+0.49$\,pp on average across all 74 (lang, hours, init)
cells, so the FP32 conclusions carry over to the INT4-deployed
model.

More broadly, these results suggest that multilingual encoder
pretraining should be viewed less as a permanent architectural
advantage and more as a data-efficiency mechanism: it changes how
quickly a target language reaches a given WER, but its effect can
be washed out by sufficient supervised target-language data.  In
this setting, latency and INT4 encoder quantization affect the
absolute operating point, but they do not substantially change
the transfer-learning decision.  This separates three choices that
are often coupled in streaming-ASR deployment discussions: use multilingual
initialization to reduce the data needed for adaptation, choose
the streaming latency from product constraints, and apply
weight-only quantization with a small measured WER budget.

\textbf{Scope and reproducibility.}  Our study uses eight European
target languages and a single
0.6\,B cache-aware FastConformer RNN-T family with a fixed
fine-tuning recipe and multi-latency training.  We expect the
qualitative conclusion---that multilingual initialization mainly
improves data efficiency rather than changing the latency
decision---to generalize beyond this model, while the exact EN--ML
cross-over points and fitted exponent are empirical properties of
this sweep.  The pretraining checkpoint and the training and
evaluation code are released publicly to facilitate independent
reproduction and reuse.

\section*{Acknowledgment}

The author used ChatGPT and Claude to assist with the
scripts for model training, evaluation and figure
rendering (Sections~\ref{sec:setup}--\ref{sec:quant}) and with
grammar editing of the manuscript.  AI was not used to
generate the experimental data, numerical results, scientific
claims, or conclusions; all code, results and figures were
reviewed, validated and finalized by the author, who takes full
responsibility for the content of this article.


\end{document}